\documentclass[a4]{article} 
\usepackage{times}

\usepackage{hyperref}
\usepackage{url}
\usepackage{eulervm}
\usepackage{charter}
\usepackage[scaled]{beramono}
\usepackage[T1]{fontenc}
\usepackage{graphicx}

\usepackage{amsmath}
\usepackage{amssymb}

\usepackage{float}

\newcommand{\p}{\mbox{\,.}}

\newcommand{\B}{{\mathbb B}}
\newcommand{\cG}{{\mathcal G}}

\newcommand{\N}{{\mathbb N}}

\newcommand{\hide}[1]{}

\title{A tutorial on MDL hypothesis testing for graph analysis}

\author{Peter Bloem \\ 
       Knowledge Representation and Reasoning Group\\
       VU University Amsterdam \\ \and
       Steven de Rooij \\ 
       Mathematical Institute \\
       University of Leiden}

\begin{document}

\maketitle

\begin{abstract}
\noindent This document provides a tutorial description of the use  of the MDL principle in complex graph analysis. We give a brief summary of the preliminary subjects, and describe the basic principle, using the example of analysing the size of the largest clique in a graph. We also provide a discussion of how to interpret the results of such an analysis, making note of several common pitfalls.
\end{abstract}

\section{Introduction}

When analysing graph structure, it can be difficult to determine whether patterns found are due to chance, or due to structural aspects of the process that generated the data. Hypothesis tests are often used to support such analyses. These allow us to make statistical inferences about which null models are responsible for the data, and they can be used as a heuristic in searching for meaningful patterns.

The minimum description length (MDL) principle \cite{rissanen1978modeling,grunwald2007minimum} allows us to build such hypothesis tests, based on efficient descriptions of the data. Broadly: we translate the regularity we are interested in into a \emph{code} for the data, and if this code describes the data more efficiently than a code corresponding to the null model, by a sufficient margin, we may reject the null model. 

This is a \emph{frequentist} approach to MDL, based on hypothesis testing. Bayesian approaches to MDL for model \emph{selection} rather than model \emph{rejection} are more common, but for the purposes of pattern analysis, a hypothesis testing approach provides a more natural fit with existing literature.

We provide a brief illustration of this principle based on the running example of analysing the size of the largest clique in a graph. We illustrate how a \emph{code} can be constructed to efficiently represent graphs with large cliques, and how the description length of the data under this code can be compared to the description length under a code corresponding to a \emph{null model} to show that the null model is highly unlikely to have generated the data.

\section{Preliminaries}

We require some basic notions from graph theory and from information theory. We describe these briefly, to define our notation, and to provide a concise introduction. For a more detailed introduction, please follow the references provided.
\paragraph{Graphs} A graph $G$ of size $n$ is a tuple $(V_G, E_G)$ containing a set of \emph{nodes} (or \emph{vertices}) $V_G$ and a set of \emph{links} (or \emph{edges}) $E_G$. For convenience in defining probability distributions on graphs, we take $V_G$ to be the set of the first $n$ natural numbers. $E_G$ contains pairs of elements from $V_G$. For the dimensions of the graph, we use the functions $n(G) = |V_G|$ and $m(G) = |E_G|$. If a graph $G$ is \emph{directed}, the pairs in $E_G$ are ordered, if it is \emph{undirected}, they are unordered. A \emph{multigraph} has the same definition as a graph, but with $E_G$ a multiset, i.e. the same link can occur more than once.

We will limit ourselves here to datasets that are (directed or undirected) \emph{simple graphs}: i.e. no link connects a node to itself and the same link cannot occur more than once. Probability distributions on the set of simple graphs are usually the most complex, so that we can trust that a method that works for simple graphs is easily translated to other types.

Two graphs $G$ and $H$ are \emph{isomorphic} if there exists a bijection $f: N_G \to N_H$ on the nodes of $G$ such that two nodes $a$ and $b$ are adjacent in $G$ if and only if $f(a)$ and $f(b)$ are adjacent in $H$. If two graphs $G$ and $H$ are isomorphic, we say that they belong to the same \emph{isomorphism class}.

\paragraph{Codes and MDL} The intuition behind MDL is that compressing data and learning from it are much the same process: in both cases, we are analyzing data to find patterns that are characteristic for the data, or patterns that represent the data well. This is not just an intuition, there exists a very precise correspondence between optimizing for probability and optimizing for description length. We will detail the basic principle below, and give an intuitive example of how we apply the method in the next section. For more details, we refer the reader to \cite{grunwald2007minimum}.
 
Let $\B$ be the set of all finite-length binary strings. We use $|b|$ to represent the length of $b \in \B$. Let $\log(x) = \log_2(x)$. A \emph{code} for a set of objects $\cal X$ (usually graphs, in our case) is an injective function $f: {\cal X} \to \B$, mapping objects to binary code words. All codes in this paper are \emph{prefix-free}: no code word is the prefix of another. We will denote a \emph{codelength function} with the letter $L$, ie. $L(x) = |f(x)|$. It is common practice to compute $L$ directly, without explicitly computing the codewords. In fact, we will adopt the convention of referring to $L$ itself as a code.

A well known result in information theory is the association between codes and probability distributions, implied by the  \emph{Kraft inequality}: for each probability distribution $p$ on $\cal X$, there exists a prefix-free code $L$ such that for all $x \in \cal X$: $- \log p(x) \leq L(x) < -\log p(x) + 1$. Inversely, for every prefix-free code $L$ for $\cal X$, there exists a probability distribution $p$ such that for all $x \in \cal X$: $p(x) = 2^{-L(x)}$. For proofs, see \cite[Section~3.2.1]{grunwald2007minimum} or \cite[Theorem~5.2.1]{cover2006elements}. To explain the intuition, note that we can easily transform a code $L$ into a sampling algorithm for $p$ by feeding the decoding function random bits until it produces an output. To transform a probability distribution to a code, techniques like arithmetic coding \cite{rissanen1979arithmetic} can be used. 

As explained in \cite[page 96]{grunwald2007minimum}, the fact that $-\log p^*(x)$ is real-valued and $L^*(x)$ is integer-valued can be safely ignored and we may \emph{identify} codes with probability distributions, allowing codes to take non-integer values. 

When we need to encode a single choice from a finite set $S$ of options, we can use the code with length $\log |S|$, corresponding to a uniform distribution on $S$. In some cases, we can allow codes with multiple codewords for a single object. If a code $L$ has multiple codewords for some object  $x$, we may indicate the choice for a particular codeword by a parameter $a$ as $L(x; a)$.

\section{Model Rejection by Codelength}

\label{section:model-selection}

\begin{figure}[tb]
  \includegraphics[width=\textwidth]{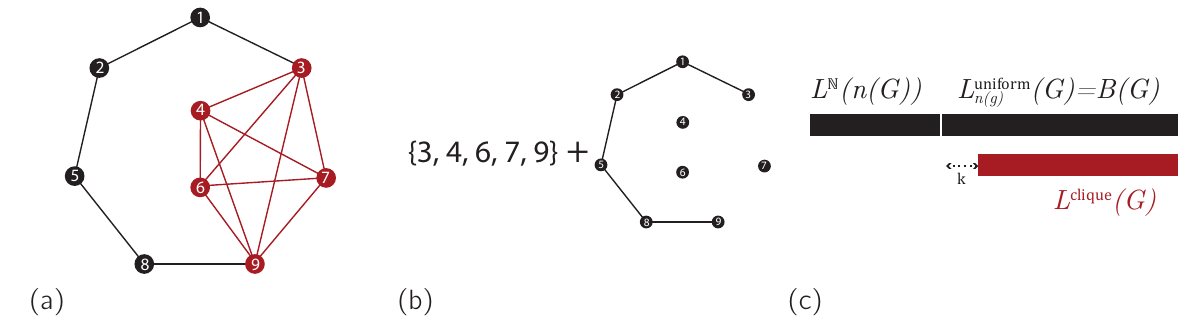}
  \caption{A simple example of graph analysis by MDL: finding cliques. (a) The data. (b) A representation of the data that exploits the existence of a large clique. We first store a set indicating which nodes belong to the clique. We then store the graph, omitting those links that belong to the clique. (c) An illustration of the principle of a bound. For a complete code on graphs, using $L^{\text{uniform}}$, we need to first store the size of the graph using some arbitrary code on the integers $L^\N$. To make sure that it does not matter which code we choose for $L^\N$, we require our code ($L^{\text{clique}}$) to beat the bound $B(G)$ (by at least $k$ bits). }
   \label{figure:clique-example}
\end{figure}  

In this section we provide an intuitive discussion of the principle of MDL hypothesis testing for graph analysis. We will illustrate the idea on a simple problem: determining whether a single large clique in a graph is an unusual feature, or to be expected under a particular null model. For certain null models, such as the Erd\H{o}s-Renyi (ER) model, the expected size of the largest clique is well understood \cite{bollobas1976cliques}. For other models, less so. Without resorting to sampling a large number of graphs from the model, how can we determine that a particular size of clique is unusually large under a given null model? We will define two codes, $L^\text{null}$ and $L^\text{clique}$, and show that if $L^\text{clique}$ compresses the data better than $L^\text{null}$ by enough bits, it is very unlikely that the data came from the null model, and we cannot dismiss the clique as a consequence of random chance.

Consider an undirected graph $G$ containing a large clique: all nodes in some subset $V_C \subseteq V_G$ are directly connected to one another. We can describe the graph by first describing the set $V_C$, and then describing $G$ in a canonical manner. Since every node in $V_C$ is connected to every other node in the clique, we can omit these links from the second part of our description, shortening the total description length, if $V_C$ is large enough. This gives us a code $L^\text{clique}$ with short codelengths for graphs containing a single large clique, and---by the correspondence mentioned in the preliminaries---also a probability distribution $p^\text{clique}$ with high probabilities for graphs with a large clique. The principle is illustrated in Figure~\ref{figure:clique-example}.

It is important to be aware that this is a code with \emph{multiple codewords for the same graph}. To convert this to a probability distribution, we sum the probabilities of all codewords: $p^\text{clique}(G) = \sum_c 2^{-L^\text{clique}(G;c)}$, where each $c$ identifies a clique in $G$.\footnote{We can convert $p^\text{clique}$ back to a another code by taking the negative logarithm: $ L^\text{clique*}(G) = -\log(G)$. $\bar L^\text{clique*}$ is a superior code (it always lowerbounds $L^\text{clique}$), but inefficient to compute. See also Section~\ref{section:multiple-testing}} As we shall see, we can use such codes without having to compute this sum, \emph{so long as they are not used as the null model}. For null models we require a code with a single codeword per graph.

Of course, there is no guarantee that, of all the distributions with a bias towards large cliques, $p^\text{clique}$ is the precise distribution from which we sampled our data. Luckily, it does not need to be: so long as we can show that our clique-based model encodes the data more efficiently than the null model, the presence of the clique disproves the hypothesis that the data came from the null model. We do not know where it came from, but it did not come from the null model. 

Under the hypothesis that the null model was the source of the data, we can show that the probability that any other model compresses the data better by $k$ bits or more, decays exponentially in $k$. This is known as  the \emph{no-hypercompression inequality} \cite[p103]{grunwald2007minimum}. Let $p^\text{null}(x)$ be any probability distribution, with $L^\text{null}(x) = - \log p^\text{null}(x)$ and let $L(x)$ be \emph{any} code, then we have
\begin{equation}
p^\text{null}\left[L^\text{null}(x) - L(x) \geq k\right] \leq 2^{-k} \p
\label{line:no-hypercompression}
\end{equation}

To provide some intuition for why this should be so, imagine that $L^\text{null}$ and $L$ are both codes for uniform distributions: $L^\text{null}$ provides each graph in $X^\text{null}$ with a codeword of constant length $n$, and $L$ provides each graph in $X$ with a codeword of constant length $n-k$. Then, $L^\text{null}$ provides enough codewords for $2^n$ graphs and, $L$ provides enough codewords for $2^{n-k}$ graphs. Thus, if a graph can be represented by both (and thus compressed by $k$ bits), its probability under the null model is at most $2^{-k}$. This is the basic principle allowing us to use compression for statistical purposes: \emph{there are exponentially fewer short codes than long ones, so compressible data is rare}.

To give a sense of scale; under the null model, the probability that $L^\text{clique}$ will compress the data better than the null model by 10 bits or more is less than one in one-thousand. For twenty bits, we get one in a million, for thirty bits, one in a billion, and so on. 

We can interpret this procedure as a hypothesis test: the difference in compression $\Delta$ between the null model and the alternative model is a statistic \cite[Example~14.2]{grunwald2007minimum}. The no-hypercompression inequality gives us a bound on the probability $p^\text{null}(\Delta \geq k)$. To reject the null model with significance level $\alpha$, we must find some code on the set of all graphs and show that it compresses the data better than the null model by $k$ bits, with $2^{-k} \leq \alpha$. Any code will do, as long as it was chosen before seeing the data. The better we design our code, the greater the power of the test.

Note that $\Delta$ is also the logarithm of the likelihood ratio between the null model and $L$, so we can see this as a likelihood ratio test. We can also interpret the difference in codelength between two models $p_a$ and $p_b$  as the logarithm of the \emph{Bayes factor} $p_a(x)/p_b(x)$ \cite[Section~14.2.3]{grunwald2007minimum}. 

Now, while our test  \emph{rejects} the null model, it does not necessarily \emph{confirm} anything about the pattern we used to compress the data. We would like to make it as likely as possible that it was the pattern we are interested in (e.g. the clique) that allowed us to reject the null model, and not some other aspect of the alternative model. This is a complex problem and it is difficult to provide any guarantees. If the alternative model stores, say, the linkset of a graph more efficiently than the null model does, it may well be that the alternative model beats the null models on all graphs in our domain, regardless of the pattern used. 

To rule out most such artifacts, we must re-use the same model we used for the null model within the alternative model to store every part of the graph except the pattern. There is usually an intuitive way to do this: for instance, in the alternative model for the clique, we store the graph in two parts: first the nodes belonging to the clique, and then the full graph, minus any links in the clique. If we use, say, the Erd\H{o}s-Renyi model as our null model, we also use it to store the graph in this second part of the code. In this way, we know that any compression achieved must be due to the clique.\footnotemark 

\footnotetext{The lack of guarantees may be a cause for skepticism. Consider, however, that a hypothesis test by itself  is \emph{never} proof that the pattern found is meaningful, only that the null hypothesis is incorrect. See also Section~\ref{section:multiple-testing}.}

\subsection{Rejecting Multiple Null Models}

\label{section:multiplerejection}

A final benefit of this method is that we can reject multiple null models with a single test. In many situations we will have a function $B(G)$ that lowerbounds any code in some set $\cal L$. If our alternative model provides a codelength below $B(G) - k_\alpha$ with $k_\alpha$ the number of bits required for our chosen $\alpha$, we can reject all of $\cal L$.

As an example, Let $\cG_n$ be the set of all undirected graphs of size $n$. For our null model, we define a uniform code on such graphs: $L^\text{uniform}_n = \log |\cG_n|$. This null model captures the idea that the size of the graph is the only informative statistic: given the size, all graphs  are equally likely. However, it is \emph{parametrized}. It is currently not a code on \emph{all} graphs, just those of size $n$. To turn it into a code that can represent all graphs, we need to encode the parameter $n$ as well, with some code $L^\N$ over the natural numbers
\[
L^\text{complete}(G) = L^\N(n(G)) + L^\text{uniform}_{n(G)}(G) \p
\]
This is called \emph{two-part coding}: we encode the parameters of a model first, and then the data given the parameters. For some parametrized model $L_\theta$, we can choose any code for $\theta$ to make it complete. We will call the set of all such complete codes the \emph{two-part codes on $L_\theta$}. Note that we can simply concatenate the two codewords, since all codes are prefix-free.

Which code we choose for the parameter is arbitrary. We may be able to reject the uniform code for one choice of $L^\N$, or several, but how can we prove that $L^\text{complete}$ will be rejected whatever $L^\N$ we choose? Instead of choosing an arbitrary code for the size, we can use the \emph{bound} $B(G) = L^\text{uniform}_{n(G)}(G)$ as our null model. This is not a code, but it \emph{is} a lower bound for any two-part code on $L^\text{uniform}_n$. If $L^\text{clique}(G)$ is shorter than $B(G)$, it is also shorter than $L^\text{complete}(G)$ whatever the choice of $L^\N$.\footnotemark

\footnotetext{
In probabilistic terms, the code on the parameter corresponds to a prior on the parameter. The two-part codes correspond to maximum likelihood posterior probabilities: $p(\hat\theta) p(x \mid\hat\theta)$. Our bound corresponds to the maximal likelihood of the data: $p(x \mid \hat \theta)$. This shows us that the bound applies not only to the two-part codes, but also to the full Bayesian mixture: $\sum_\theta p(x \mid \theta) p(\theta) \leq \sum_\theta p(x \mid \hat\theta) p(\theta) = p(x \mid\hat\theta)$
} 


Note that when we store the rest of the graph within $L^\text{clique}$ we \emph{cannot} use $B(G)$ in place of $L^\text{complete}(G)$. We want a \emph{conservative} hypothesis test: the probability of incorrectly rejecting a null model may be lower than $\alpha$ but never higher. By this principle, bounds chosen in place of either model should always decrease $\Delta$. The code corresponding to the null model must always be lowerbounded, and the optimal code for the alternative model must always be upperbounded.

This is also the reason that we allow the alternative code to have multiple codewords for one graph. Replacing such a code with the complete equivalent model, found by summing over all codewords, would increase $\Delta$, strengthening the hypothesis test. By evaluating only one codeword, we are effectively using an upper bound on this complete model, trading off power for efficiency of computation.

\section{Discussion}

We will conclude by discussing several subtleties n the interpretation of this method.

\subsection{Multiple Testing and Significance as Pattern Evidence}
\label{section:multiple-testing}

In some cases, we may test multiple patterns for the same graph. For instance, we may test several cliques found in the same data, to strengthen our attempt to reject the null hypothesis. At first glance this may seem like multiple testing, raising the question of whether we should adjust our rejection region. Does the probability that one or more of the cliques we find to be significantly large is a ``false positive'' increase with the number of subgraphs checked?

This question highlights an important issue in the use of hypothesis testing in pattern analysis: the question about the pattern we are trying to answer---is $C$ a meaningful clique?---is not the question we are testing for. Strictly interpreted, the hypothesis test investigates the question \emph{whether the full data came from the null model}. Thus, while we test multiple cliques, we only test them on a \emph{single} null-hypothesis. To perform a valid single null-hypothesis test, we can simply select the best compressing clique out of all the cliques tested, and use the log-factor of that clique as the significance with which we reject the hypothesis that the null model generated the data. 

This is a valid code: we can check as many patterns as we like before choosing the one we will use to compress the data with. Since it is a valid code, it is a valid, \emph{single} hypothesis test, and there is no need to correct for multiple testing.\footnote{If we test multiple \emph{null models} on a single dataset, multiple testing \emph{should} be taken in to account.}

To make this more precise, assume that we have some data $G$. The null model assigns codelength $L^\text{null}(G)$ and the clique code assigns $L^\text{clique}(G; C)$ for a particular clique $C$. 
The clique code gives us many different ways to describe this graph (one code word for each clique). To remove this inefficiency, we can define a code that sums the probability mass of all these codewords, so that it has only a single codeword per graph:
\begin{align*}
p^\text{clique}(G; C) &= \text{exp}_2 \left [ {- L^\text{clique}(G; C)}\right] \\
p^\text{clique*}(G) &= \sum_{C} p^\text{clique}(G; C) \\
L^\text{clique*}(G) &= - \log p^\text{clique*}(G)
\end{align*}
The codeword that $L^\text{clique*}$ provides for $G$ is smaller than any codeword provided by $L^\text{clique}$.

We can now use $L^\text{clique*}$ to perform a \emph{single} test: if $L^\text{clique*}$ compresses better than $L^\text{null}$, then by the no-hypercompression inequality (\ref{line:no-hypercompression}), we can reject the null model. We cannot compute $L^\text{clique*}$, for any reasonable size of graph, but we \emph{can} upperbound it. If we can find only \emph{one} codeword that allows $L^\text{clique}$ to compress better than $L^\text{null}$ by the required margin, we have proved that $L^\text{clique*}$ does so as well.\footnotemark~This allows us to test as many cliques as we like, and we may use exhaustive search or sampling to find these: the real test is whether $L^\text{clique*}$ compresses the data better than the null model.

\footnotetext{As usual, if we do not find such a codeword, we do not know whether or not the null hypothesis can be rejected.}

Of course, in general practice, we are not actually interested in whether or not the null model is true. What we are investigating is whether a given \emph{pattern} is ``true''. We are using the framework of hypothesis testing merely as a \emph{heuristic} to guide unsupervised pattern mining. Does the general intuition of multiple hypothesis testing still apply? Depending on the type of pattern under investigation, we may not be able to exclude artifacts similar to false positives: every subgraph of the largest clique may prove a ``positive'' pattern, even though it is not the the pattern we are ultimately looking for.

However, increasing the number of patterns checked does not raise the amount of such false positives. This is because we analyze multiple patternss on a \emph{single dataset}. The process generating the data is stochastic, but once the data is known, the log-factors of all patterns are determined. Checking more patterns will get us closer to the true number of positives, but it will not put us in danger of an increased number of \emph{false} positives.

\subsection{Limitations and Extensions}

Whether the approach described here is useful for a particular type of pattern, depends on its properties. One illustrative example is an  \emph{anti-motif}: a subgraph that occurs less than expected under the null model. In principle, any deviation from the null-model can be used to compress the data. However, computing such a code efficiently for anti-motifs may be a challenging task. For other patterns (hub nodes, motifs, $k$-regularity), suitable codes are easier to devise.

Note also that the constraints on the null model are stronger than those on the alternative model. In our code, we technically use the length of one of many codewords, rather than the true probability (which would require a sum over all codewords). In the null model, such approximations are \emph{not} appropriate, and each graph must have a single codeword. For some models (such as the configuration model \cite{newman2010networks}), this makes the codelength so expensive to compute that we may gain little over traditional methods of hypothesis testing. 

Many null models can easily be used in this framework. For instance, for the $G(n, m)$ and $G(n,p)$ models can  the codelength can easily be computed. For other models, like the preferential attachment model \cite{albert2002statistical}, we can easily come up with a code with multiple codewords for each graph, but a code with a single codeword per graph is more complex to compute efficiently.

 \paragraph{Acknowledgements} This publication was supported by the Dutch national program COMMIT, by the Netherlands eScience center, and by the Amsterdam Academic Alliance Data Science (AAA-DS) Program Award to the UvA and VU Universities. 
 
\bibliographystyle{plain}
\bibliography{pattern.bib}
	
\end{document}